\documentclass[letterpaper]{article} 
\usepackage{aaai24}  
\usepackage{times}  
\usepackage{helvet}  
\usepackage{courier}  
\usepackage[hyphens]{url}  
\usepackage{graphicx} 
\usepackage{natbib}  
\usepackage{caption} 
\usepackage{url}
\usepackage{amsmath}
\usepackage{amssymb}

\usepackage[ruled,vlined]{algorithm2e}
\usepackage{booktabs}
\usepackage{color}
\usepackage[table,xcdraw]{xcolor}
\usepackage{multirow}
\usepackage{algorithmic}
\title{Test-Time Personalization with Meta Prompt for Gaze Estimation}
\author{
    Huan Liu\textsuperscript{\rm 1}, Julia Qi\textsuperscript{\rm 1,\rm 2}\equalcontrib\thanks{Work done during an internship at Huawei Noah's Ark Lab.}, Zhenhao Li\textsuperscript{\rm 1}\equalcontrib, Mohammad Hassanpour\textsuperscript{\rm 1}, Yang Wang\textsuperscript{\rm 3}, \\
    Konstantinos Plataniotis\textsuperscript{\rm 4}, Yuanhao Yu\textsuperscript{\rm 1}
}
\affiliations{
    \textsuperscript{\rm 1}  Noah's Ark Lab, Huawei, Canada \\
    \textsuperscript{\rm 2} University of Waterloo, Canada\\
    \textsuperscript{\rm 3} Department of Computer Science and Software Engineering, Concordia University, Canada\\
    \textsuperscript{\rm 4} The Edward S. Rogers Sr. Department of Electrical and Computer Engineering, University of Toronto, Canada\\
    \{huan.liu127, zhenhao.li1, mohammad.hassanpour, yuanhao.yu\}@huawei.com, j6qi@waterloo.ca, \\
    yang.wang@concordia.ca, 
kostas@ece.utoronto.ca
    
}

\usepackage{bibentry}

\begin{document}

\maketitle

\begin{abstract}
Despite the recent remarkable achievement in gaze estimation, efficient and accurate personalization of gaze estimation without labels is a practical problem but rarely touched on in the literature. 
To achieve efficient  personalization, we take inspiration from the recent advances in Natural Language Processing (NLP) by updating a negligible number of parameters, ``prompts", at the test time. 
Specifically, the prompt is additionally attached  without perturbing original network and can contain less than 1\% of a ResNet-18's parameters. Our experiments show high efficiency of the prompt tuning approach. The proposed one can be 10 times faster in terms of adaptation speed than the methods compared. 
However, it is non-trivial to update the prompt for personalized gaze estimation without labels.  At the test time, it is essential to ensure that the minimizing of particular unsupervised loss leads to the goals of minimizing gaze estimation error. To address this difficulty, we propose to meta-learn the prompt to ensure that its updates align with the goal. Our experiments show that the meta-learned prompt can be effectively adapted even with a simple symmetry loss. In addition, we experiment on four cross-dataset validations to show the remarkable advantages of the proposed method. Code is available at \url{https://github.com/hmarkamcan/TPGaze}.
\end{abstract}
\section{Introduction}

Gaze, which refers to the direction of an individual's visual focus, is a crucial indicator of human attention. 
Gaze estimation is a rapidly evolving area of research, with the aim of determining the direction of an individual's gaze based on the positioning and orientation of their eyes and head. Due to its potential applications in diverse fields such as healthcare \cite{med}, gaming \cite{game1}, and human-computer interaction \cite{hci1}, it has attracted significant attention.

\begin{figure}[t]
\centering
\includegraphics[width=1.0\linewidth]{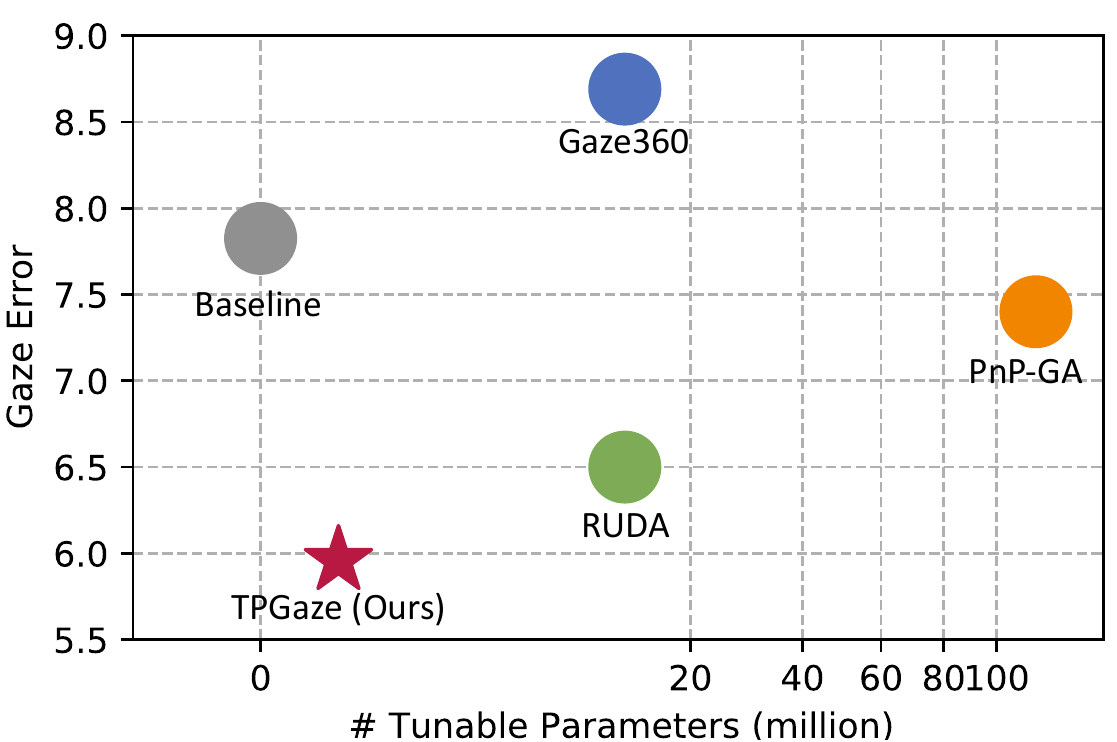}
\caption{Illustration of performance comparison in terms of gaze error and tunable parameters. Our method achieved the lowest gaze error with negligible tunable parameters required for personalization. Gaze errors are calculated by averaging four cross-dataset validations.}
\label{fig:first}
\end{figure} 

Recent years have witnessed tremendous success of utilizing deep learning in addressing the gaze estimation problem. However, most of deep learning based methods \cite{eye-everyone,rt-gene,gaze360,eyediap,xgaze,mpiigaze} essentially learn a mapping on the training data in a supervised manner. Although achieving high accuracy on the training dataset, these methods suffer from performance degradation when tested in real-world scenarios with distribution shift. In addition, collecting labeled data in the real world is extremely difficult, making it challenging to fine-tune. These issues raise concerns about the practical value of purely supervised methods. To address the above concern, recent attempts have focused on gaze estimation in the unsupervised domain adaptation (UDA) setting \cite{gaze360,wang2019generalizing, pnp-gaze}. However, these UDA methods assume the availability of source data, which may not be available in real-world applications due to privacy concerns. In the literature, only few methods, such as  \cite{CRGA} and \cite{RUDA}, consider addressing UDA without source data.

In this paper, we study a new variant of UDA, i.e., 
test-time
personalization. With the widespread use of portable devices such as smartphones, laptops and tablets, personalized experiences have become increasingly necessary. Personalizing gaze estimation based on the user's unique characteristics is essential for enhancing their experience. Unfortunately, current personalized gaze estimation methods usually require either calibration \cite{eye-everyone} or ground truth gaze labels \cite{few-shot-gaze, differential, persona20, persona19, mtgls}, which may significantly limit their practical deployment. One possible solution is to directly employ \cite{CRGA} and \cite{RUDA} for the adaptation on personal data. However, applying these methods under test-time personalization settings is non-trivial. They require all parameters to be trainable during the adaptation phase, which may not be practical to be conducted on edge devices at the test time due to computational constraints. 

To this end, we propose a \textbf{T}est-time \textbf{P}ersonalized \textbf{Gaze} estimation (\textbf{TPGaze}) method by considering both adaptation efficiency and effectiveness.
Specifically, we take inspiration from natural language processing research \cite{nlp1, nlp2,nlp3}  and propose to update a small group of parameters solely, namely the ``prompt'', while freezing the backbone of the network during personalization. The prompt is person-specific and memory-saving, with a cost of less than 1\% of a ResNet-18 model.
Ideally, it might be feasible to update the prompt for performance improvements using unsupervised gaze-relevant losses, such as the rotation consistency loss \cite{RUDA}.
These loss functions are carefully yet intuitively designed to be correlated to gaze.

The effectiveness of the losses is primarily demonstrated through empirical studies, such as experimental results.
However, it is somewhat hard to guarantee that the gradients with respect to such losses align with the direction that minimizes gaze estimation error for any particular person. 
To bridge unsupervised losses and gaze error, 
we propose a meta-learning-based approach that can explicitly associate the two objectives. The goal of meta-learning here is to learn an ideal initialization of the prompt across individuals so that its updates towards lower unsupervised losses are equivalent to updates towards lower gaze estimation error. We will show that meta-learning is very effective even with a simple left-right symmetry loss \cite{gaze360}.

Our main contributions are summarized as follows:
1) We propose an efficient method for test-time personalized gaze estimation that achieves fast adaptation by leveraging prompt. 
2) Our proposed method employs meta-learning to initialize the prompt, explicitly ensuring that test-time prompt updates result in reduced gaze estimation error. 
3) Extensive experimental results demonstrate the effectiveness of our proposed method. Our approach not only significantly outperforms existing methods but also adapts at least 10 times faster than the compared methods. To illustrate, Figure \ref{fig:first} presents a preview of the superior performance of our method. 
\section{Related Work}

\subsection{Appearance-based Gaze Estimation}





Research on gaze-direction-from-eye-appearance has been ongoing for over a century \cite{paper1824}. Early approaches  \cite{eyetab, eyeball10} mainly rely on constructing geometric eye models from images.
In the recent decade, deep learning on labeled datasets \cite{mpiigaze, eye-everyone, mpiigazeface, eyediap, gaze360, xgaze} has been a game changer to the field, eliminating the need to explicitly construct eye models. Deep neural network (DNN) models present a solid ability to learn rich gaze-relevant features from supervised learning, achieving breakthroughs in estimation accuracy \cite{eye-everyone, arenet, dpg, farenet, canet, agenet, lian2018multiview,lian2019rgbd}. Besides, GazeNeRF \cite{gazenerf} propose a 3D-aware design by incorporating neural radiance fields to synthesize more samples for effective supervised training.


Transfer learning techniques, such as domain adaptation, have been adopted in recent gaze estimation approaches to improve the performance of a DNN model across datasets.
Most follow the UDA setting \cite{gaze360, wang2019generalizing, pnp-gaze, RUDA, DAGEN}, in which labeled source domain data and unlabeled target domain data are accessible during adaptation. 
CRGA \cite{CRGA}, RUDA \cite{RUDA} and UnReGA\cite{cvpr-sfgaze} perform source-free UDA without the need of source domain data, though still requiring a large amount of target domain data.
Another setting is few-shot personalization, which utilizes a few labeled data samples to optimize the model performance for a specific person \cite{differential, persona19, persona20, few-shot-gaze}. However, their experiments rely on gaze labels from off-the-shelf datasets, neglecting the difficulty to obtain such labels in practice. In this paper, we propose to address a more challenging problem, i.e., unsupervised personalization without source data. 

\subsection{Prompt Tuning for Computer Vision}

Large foundation models have demonstrated exceptional
effectiveness in natural language processing (NLP) and
computer vision tasks. 
To fast adapt large models to downstream tasks, researchers have proposed continuous task-
specific vectors that are updated via gradient, a method
known as prompt tuning \cite{nlp3, nlp2, liu2021p}.
In prompt tuning, the backbone parameters are fixed and only prompts are updated. As a parameter-efficient method, prompt tuning can achieve comparable results to full-parameter fine-tuning.  
In computer vision, prompting has initially been introduced to generalizing language models to address the few-shot and zero-shot classification problems \cite{clip}. Despite its effectiveness, most works \cite{clip,vl1,vl2,vl3} use both vision and language models and implement prompt by applying it only to the language model. Recently, \cite{vpt} initially attempted to apply prompt tuning on pure vision models for few-shot learning. In \cite{vpt}, the authors propose two variants of prompt, i.e., additional inputs of vision transformer or tunable padding of convolutional layers. Following this work, \cite{padding,padding2} propose to use the prompting method in \cite{vpt} for speaker-adaptive speech recognition. In this paper, we follow \cite{vpt} to apply tunable padding as prompt. In addition, we move one step forward to learn a meta-initialized prompt for better generalizing to a specific person’s data.

\subsection{Meta-Learning}
Meta-learning, also known as learning-to-learn, has been widely applied in deep learning \cite{meta_model1,meta_metric1,maml,NEURIPS2022_8bd4f1db,lian2019towards}. Meta-learning methods can be categorized as model-based \cite{meta_model1}, metric-based \cite{meta_metric1} and optimization-based \cite{maml}. An example is the model-agnostic meta-learning (MAML) \cite{maml} method. It aims to jointly learn a global optimal initialization of parameters and an update rule of task-specific
parameters. As a result, the meta-learned model can generalize well to target tasks given a small number of samples with few gradient updates. Considering its ability to enable few-shot learning,  \cite{few-shot-gaze} propose to use meta-learning for few-shot gaze estimation. It proposes to use meta-learning to learn how to quickly adapt to unseen domains with few labeled data. Our use of meta-learning is different than that in \cite{few-shot-gaze}. Instead, our meta-learning approach is closer to a variant of MAML, i.e., meta-auxiliary learning \cite{liu2019self, chi2021test, Liu_2022_CVPR,Liu_2023_WACV}. It aims to address the primary task by solving an auxiliary task. 
\section{Problem Definition}\label{sec:prob_def}
In this paper, we focus on a rarely touched problem, i.e., test-time personalized gaze estimation. In contrast to unsupervised domain adaptation (UDA) for gaze estimation \cite{pnp-gaze, CRGA, RUDA}, which emphasizes the average performance across all persons in the target domain, test-time personalization focuses on the performance with respect to a particular person in the target domain. To be specific, let $\mathcal{S}=\{(x_i^s,y_i^s)|x_i^s\in \mathcal{I}_S,y_i^s\in \mathcal{Y}_S\}_{i=1}^{N_S}$ denote the source dataset, where $x_i^s$ and $y_i^s$ are respectively the image and label from source image set $\mathcal{I}_S$ and source label set $\mathcal{Y}_S$. Similarly, let $\mathcal{A}_j=\{x_i^{a_j}|x_i^{a_j}\in \mathcal{I}_{T_j}\}_{i=1}^{N_{\mathcal{A}_j}}$ denote the personalization dataset of $j$-th person, where $x_i^{a_j}$ is the unlabeled image sampled from the target image set of $j$-th person $\mathcal{I}_{T_j}$. 
Our goal is to update the model $f_\theta$ learned on the source dataset $S$ according to the personalization dataset $\mathcal{A}_j$,
so that the resulting personalized model $f_{\theta_{j}}$ can perform better on the test data of the $j$-th person. 

\section{Preliminary of Source-Free UDA} \label{sec:preliminary}

\begin{figure}[!t]
\centering
\includegraphics[width=\linewidth]{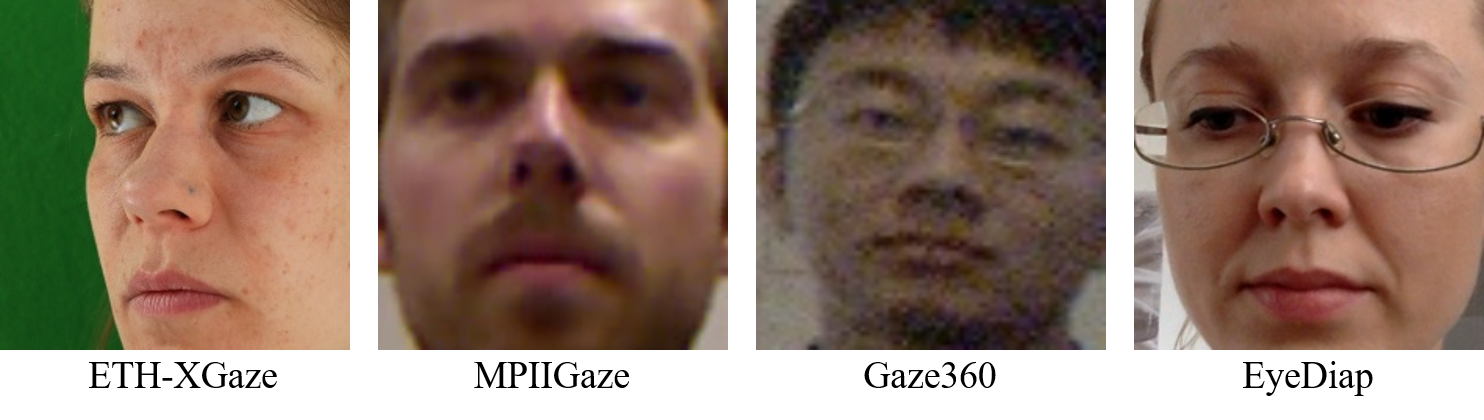}
\caption{Illustration of the difference between four representative datasets. They are different from each other in subject appearance, image quality and lighting conditions. }
\label{fig:dataset}
\end{figure}

We begin with the formulation of source-free unsupervised domain adaptation (UDA) before introducing our proposed approach. 
UDA methods \cite{RUDA,pnp-gaze} consider a realistic problem that a model pre-trained on a dataset should be generalized to unseen data with distribution shift due to variations in subject appearance, lighting conditions, and image quality. We illustrate such variations in Figure \ref{fig:dataset} with examples from four datasets. 

UDA methods usually start from supervised pre-training on a source dataset $\mathcal{S}$.
In the pre-training of gaze estimation on the source data, a typical practice \cite{RUDA,pnp-gaze} is to learn a mapping $f_\theta$
via a combination of supervised and unsupervised losses:

\begin{equation} \label{eq:pre-train_init}
    \theta = \arg\min_{\theta} (\mathcal{L}_1(f_\theta(\mathcal{I}_S), \mathcal{Y}_S) + {\mathcal{L}}_{un}(f_\theta(\mathcal{I}_S), \mathcal{I}_S)),
\end{equation}

\noindent
where $\mathcal{L}_{1}$ is the supervised loss and $\mathcal{L}_{un}$ denotes the unsupervised loss. This pre-training phase aims to learn a model that can estimate gaze and accomplish unsupervised tasks, such as maintaining rotation consistency \cite{RUDA}.

At the adaptation stage, only images $\mathcal{I}_{T}$ of the target domain are available for adapting the gaze estimation model.
The domain adaptation of a typical UDA method, including \cite{pnp-gaze, RUDA}, can be summarized as:

\begin{equation} \label{eq:pre-adaptation}
    \theta_j = \arg\min_{\theta}\mathcal{L}_{un}(f_\theta(\mathcal{I}_{T}), \mathcal{I}_{T}).
\end{equation}
where the model parameters $\theta$ are fine-tuned by minimizing the unsupervised loss $\mathcal{L}_{un}$.

However, we argue that such a solution is not optimal in the context of personalizing a gaze estimation model where the target domain data from one person might be limited and similar among samples.
First, optimizing a large number of parameters $\theta$ on such a small amount of data with similar patterns can easily lead to overfitting problem \cite{schneider2021personalization}.
Second, fine-tuning the entire network can be computationally costly, especially in scenarios with limited computing resources, such as adaptation on personal edge devices.
In addition, although the unsupervised loss ($\mathcal{L}_{un}$) in domain adaptation has also been employed in the pre-training phase (see Equation \ref{eq:pre-train_init}), there is no explicit constraint ensuring 
that minimizing $\mathcal{L}_{un}$ is equivalent to minimizing $\mathcal{L}_{1}$. 
This can pose a risk of unsuccessful adaptation unless $\mathcal{L}_{un}$ is very carefully designed \cite{RUDA,pnp-gaze}.

\section{Methodology}

\begin{figure}[!t]
\centering
\includegraphics[width=0.75\linewidth]{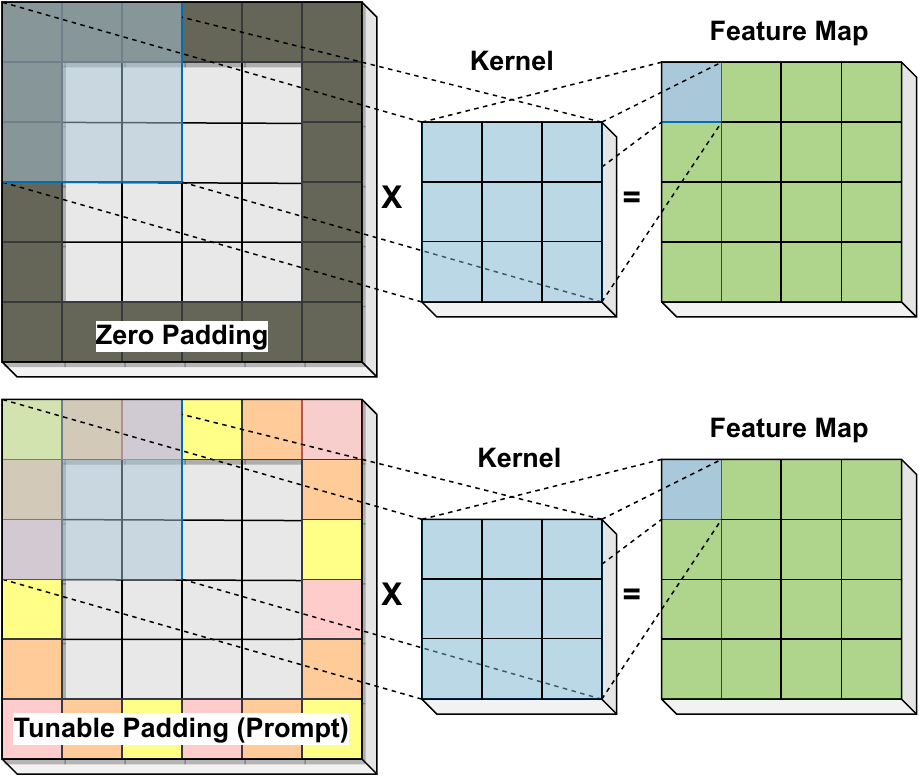}
\caption{Illustration of replacing padding by tunable prompt.}
\label{fig:prompt}
\end{figure}

\subsection{Prompting for Test-time  Personalization}\label{sec:prompt}
Unlike common machine learning approaches that learn a generic model for a broad user base, personalization acknowledges individual user characteristics, delivering a dedicated model for each. 
Test-time adaptation \cite{sun2020test} is an potential way to enable personalization during testing with limited users' data.
However, achieving test-time personalization without calibration or labels is still challenging. 

\subsubsection{Prompting for Personalized Gaze Estimation}
Inspired by the recent advances in NLP \cite{nlp1, nlp2,nlp3}, we incorporate the essence of the prompt tuning method into gaze estimation for efficient personalization. 
The key idea of prompting is to modify the inputs rather than network parameters \cite{nlp1}. Following \cite{vpt, padding}, we propose to transform the padding, an input to convolutional layers, to be our prompt.  In convolutional layers, padding is usually employed to maintain or control the size of the output feature maps, such as zero padding and reflect padding. In these paddings, a predetermined number of zeros or reflections of the inputs are attached to the border of inputs before convolution operation. As shown in Figure \ref{fig:prompt}, the padded region is then convolved with the kernel to produce the output feature map.
In this way, a change in padding can impact the resulting feature embedding.
Therefore, a natural idea is that we can modify the padding adaptively,
guiding a network to produce desired feature embedding with respect to any specific person.

\begin{figure}[!t]
\centering
\includegraphics[width=0.9\linewidth]{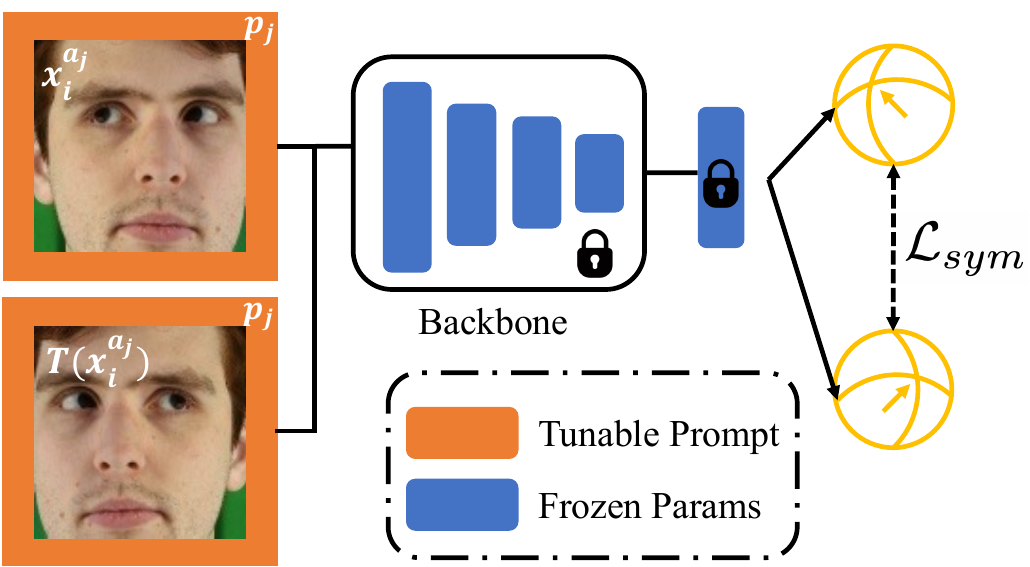}
\caption{An overview of the proposed test-time personalization on $j$-th subject. In personalization, all the parameters are fixed except for the prompt.}
\label{fig:network}
\end{figure}

To instantiate this, we replace conventional padding with tunable parameters and update them during the personalization phase. By making the padding parameters trainable, we enable the network to learn the most appropriate padding strategy for a given subject. This, in essence, enables the network to be personalized. 

It is worth mentioning that the number of parameters in prompt is far less than that in an entire network. For example, in ResNet-18, the prompt can contain less than 1\% of the model parameters.
The lightweight nature of the prompt makes it exceptionally suitable for test-time personalization. This can address the concerns raised in Section \ref{sec:preliminary} about the overfitting problem and limited computational resources on edge devices.

\subsubsection{Test-time Adaptive Gaze Estimation}
Given a model $f_\theta$ pre-trained on source data using Equation \ref{eq:pre-train_init}, we modify the model by incorporating a prompt (denoted as $p$), resulting in a new model $f_{\{\theta, p\}}$.
At the test time, as discussed above, we update the prompt $p$ instead of the entire set of model parameters practiced in existing methods \cite{pnp-gaze, RUDA}.
We formalize the test-time training by rewriting the optimization problem in Equation \ref{eq:pre-adaptation} as follows:
\begin{equation} \label{eq:prompt}
\begin{aligned}
    p_j &= \arg\min_{p}\mathcal{L}_{per}(f_{\{\theta, p\}}(\mathcal{A}_j), \mathcal{A}_j).
\end{aligned}
\end{equation}
\noindent
where $p_j$ denotes the prompt corresponding to $j$-th person. $\mathcal{L}_{per}$ is an unsupervised loss  used in both pre-training and personalization. 

For the choice of $\mathcal{L}_{per}$, we want it to be gaze-relevant and 
computationally efficient.
One option is the rotation consistency loss proposed in \cite{RUDA}. Despite its effectiveness, the calculation of this loss demands multiple rotated versions of the original image as input to the network, 
resulting in great time cost in model adaptation.
For ease of implementation and fast calculation, we adopt the left-right symmetry loss cited in \cite{gaze360}.
During the test time,
it can be applied by the pre-trained model to regularize the gaze estimates for the specific individual being tracked.
For a detailed description of the symmetry loss, please refer to the appendix. An overview of our test-time personalization is depicted in Figure \ref{fig:network}.




\subsection{Meta Prompt}\label{sec:meta}
\begin{algorithm}[t]
\caption{Training of Meta Prompt}
    \SetKwInOut{Require}{Require}
    \SetKwInOut{Output}{Output}
    \Require{ Pre-trained network: $f_{\theta}$ }
    \Require{ Prompt: $p$ }
    \Require{ learning rate $\lambda_1$ and $\lambda_2$ }
    \Output{meta-initialized prompt $p$ }
    \vspace{5pt}

    \textbf{Initialize}: Initialize the prompt $p$ in $f_{\theta}$ by Gaussian noise and allow it to be trainable.
    
    \While{not converge}{
        Sample a mini-batch of training data in $\{\mathcal{I}_S, \mathcal{Y}_S\}$;
        
        \For{each $x^s_i$}{
        
        Compute updated prompt $\hat{p}$:
        
        $\hat{p} = p - \lambda_1 \nabla_{p} \mathcal{L}_{per}(f_{\{\theta, p\}}(x^s_i), x^s_i)$
        }
        
    \textbf{Update:}
    
        $p \leftarrow p - \lambda_2 
\nabla_{p} \mathcal{L}_{1}(f_{\{\theta, \hat{p}\}}(x^s_i), y^s_i)$}

    \label{al:algorithm}
\end{algorithm}

As mentioned in Section \ref{sec:preliminary}, it is questionable that minimizing unsupervised losses $\mathcal{L}_{per}$ can naturally lead to a smaller gaze error, which is the ultimate goal.
In this section, we present how meta-learning can align these two objectives.
The recent method described in \cite{few-shot-gaze} leverages meta-learning to enable rapid adaptation of gaze estimation model provided labels from the target domain.
In this work, we explore the feasibility of constructing a ``meta prompt" that can rapidly adapt to individual characteristics without the need for gaze labels. Inspired by the recent attempts in meta-learning \cite{chi2021test, liu2019self, Liu_2022_CVPR} 
for image deblurring and classification, where the adaptation is performed via an auxiliary loss, we are further motivated to meta-initialize the prompt before tuning.     The goal of the meta-training is to learn the prompt so that the gaze error is spontaneously minimized by optimizing the prompt based on the symmetry loss.

Given a pre-trained model $f_{\theta}$, we randomly initialize the prompt $p$ and obtain the model $f_{\{\theta, p\}}$ with the tunable prompt equipped. Note that the meta-training of prompt is conducted only on the source dataset $\mathcal{S}$. 
Next, we sample a mini-batch of $N$ paired data ${\{x^s_i, y^s_i\}}^N_{i=1}$ and proceed to update the prompt for each sample using the personalization loss, as follows:

\begin{equation}\label{meta-inner}
    \hat{p} = p - \lambda_1 \nabla_{p} \mathcal{L}_{per}(f_{\{\theta, p\}}(x^s_i), x^s_i),
\end{equation}

\noindent
where $\lambda_1$ is the learning rate. Intuitively,  this update allows the prompt to guide the network in generating features that reduce the left-right symmetry loss. 

Recall that, our primary objective is to minimize the gaze error through updates to the prompt $p$. Accordingly, we aim to maximize the performance of the network by minimizing $\mathcal{L}_{per}$. To this end, we formally define our meta-objective as follows:
\begin{equation}
\arg\min_{p} \mathcal{L}_1(f_{\{\theta, \hat{p}\}}(x^s_i), y^s_i).
\label{eq:meta-objective}
\end{equation}

Note that the supervised loss is computed based on the network's result $f_{\{\theta, \hat{p}\}}(x^s_i)$, which is based on the updated prompt $ \hat{p}$.
However, the actual optimization is carried out on the prompt $p$. The meta-objective can be implemented by the following gradient descent:

\begin{equation}
\begin{aligned}
p \leftarrow p - \lambda_2 
\nabla_{p} \mathcal{L}_{1}(f_{\{\theta, \hat{p}\}}(x^s_i), y^s_i),
\end{aligned}
\label{eq:meta-outer}
\end{equation}

\noindent
$\lambda_2$ being the learning rate. 

As a summary, the above meta-training is designed to learn a general initialization of the prompt $p$ that can be applied across individuals. Notably, since the subject being tested is unknown during the meta-learning phase, it is impossible to learn a personalized meta prompt $p_j$. The overall meta-learning procedure is summarized in Algorithm \ref{al:algorithm}.

\section{Experiments}

 
\subsection{Dataset}
We employ four gaze estimation datasets as four different domains, namely ETH-XGaze ($\mathcal{D}_E$) \cite{xgaze}, Gaze360 ($\mathcal{D}_G$) \cite{gaze360}, MPIIGaze ($\mathcal{D}_M$) \cite{mpiigaze}, and EyeDiap ($\mathcal{D}_D$) \cite{eyediap}. For preprocessing, we use the code provided in \cite{Cheng2021Survey}. $\mathcal{D}_E$ and $\mathcal{D}_G$ are used as source domains whereas $\mathcal{D}_M$ and $\mathcal{D}_D$ are used as target domains, in alignment with RUDA \cite{RUDA} and PnP-GA \cite{pnp-gaze}.
Due to the page limit, please refer to supplementary material for the details of the  datasets.
\subsection{Implementation Details} \label{sec:implementation}
Our method is implemented using PyTorch library \cite{paszke2019pytorch} and conducted on NVIDIA Tesla V100 GPUs.
We use Adam \cite{kingma2014adam} as our optimizer with $\beta = (0.5, 0.95)$. The training images are all cropped to a size of $224 \times 224$ without data augmentation.

\noindent
\textbf{Pre-training stage.}
During network pre-training, we use L1 loss and symmetry loss to train the network $f_\theta$ with a mini-batch size of 120. The initial learning rate is set to $10^{-4}$. We train for 50 epochs with the learning rate multiplied by 0.1 at Epoch 25.

\noindent
\textbf{Meta-training stage.}
During the meta-training stage for prompt initialization, the network is initialized with weights obtained from the pre-training stage. Prompts are initialized randomly using a Gaussian distribution with mean 0 and variance 1. Note that if not explicitly specified, we replace the padding of the first nine convolutional layers in ResNet-18 \cite{resent}. All other parameters are kept frozen. We use a mini-batch size of 20. The learning rates, i.e., $\lambda_1$ and $\lambda_2$ in Algorithm \ref{al:algorithm}, are set to $10^{-4}$. The meta-training process continues for 1000 iterations.

\noindent
\textbf{Personalization stage.}
To perform personalization for solving Equation \ref{eq:prompt}, we use only 5 images per person. To ensure the reproducibility of results, we do not perform random sampling but use the first 5 images of each person. During personalization, only the prompt is optimized, and all other parameters are fixed. The learning rate is set to 0.01.

\subsection{Comparison with the SOTA}

To demonstrate the effectiveness and efficiency of our proposed method, we conduct a comparative study with several UDA methods for gaze estimation across four cross-domain tasks: $\mathcal{D}_E \rightarrow \mathcal{D}_M$, $\mathcal{D}_E \rightarrow \mathcal{D}_D$, $\mathcal{D}_G \rightarrow \mathcal{D}_M$, and $\mathcal{D}_G \rightarrow \mathcal{D}_D$.  We select five representative methods, including source-available UDA (SA-UDA) and source-free UDA (SF-UDA) methods. Specifically, we compare our method with three SA-UDA methods, i.e., DAGEN \cite{DAGEN}, GazeAdv \cite{wang2019generalizing} and Gaze360 \cite{gaze360}. For SF-UDA methods, we select PnP-GA \cite{pnp-gaze} and RUDA \cite{RUDA} for comparison. As SF-UDA methods can be easily adapted to our setting, we re-implemented these methods under our setting. For the re-implementations, we rely on their official code provided by the authors. For a fair comparison, the personalization of PnP-GA \cite{pnp-gaze} and RUDA \cite{RUDA} uses 10 images per person as the two methods are not specifically designed for personalization. ResNet-18 is used as the backbone for all methods.

\subsubsection{Effectiveness of Our Method} 
We present a comparison between the aforementioned methods in terms of gaze error in Table \ref{tab:comp}.
We begin by comparing our method (TPGaze) with a baseline 
 based on supervised learning on the source dataset with L1 loss.
Apparently, our method significantly outperforms the baseline in the cross-domain scenario.

Next, we compare our method with the SA-UDA approaches. All three SA-UDA approaches require labeled source data during the adaptation stage. Intuitively, these approaches can potentially achieve better performance due to the availability of source data. However, our method, which uses 5 personal images for adaptation only, significantly outperforms these SA-UDA methods by a large margin. 

Finally, we compare with two SF-UDA methods. As previously noted, we re-implement these methods under our personalization setting. The last three rows of Table \ref{tab:comp} clearly demonstrate that our method surpasses both PnP-GA and RUDA, even when those two methods are trained with 10 personal samples and employ unsupervised losses that are more complex than the symmetry loss we used. 
There can be two reasons:
1) Both PnP-GA and RUDA require all parameters to be tunable during the personalization stage. However, due to the limited amount of test domain data, this may potentially result in overfitting.
In contrast, our method only allows a very small group of parameters to be tunable, mitigating the risk to overfit.
2) PnP-GA and RUDA  assume that minimizing the outlier-guided loss or rotation consistency loss can be beneficial in reducing gaze error unconditionally. In comparison, our method proposes to use meta-learning to align the unsupervised loss with gaze error.

\subsubsection{Efficiency of Our Method}

As stated in Section \ref{sec:prompt}, 
efficient adaptation is a key advantage of our method. To prove this, we exhibit the number of tunable parameters for each method in the second column of Table \ref{tab:efficiency}. Our method requires only 0.125M parameters to be trainable during personalization, while RUDA requires at least 100 times more parameters to be updated. 
Additionally, special attention should be paid to PnP-GA, which relies on an ensemble of 10 networks and requires 116.9M parameters to be optimized for adaptation. This is 1000 times larger than our method. 
The excessive number of tunable parameters in the compared methods can potentially hinder their deployment on edge devices for online adaptation, while our method can be easily deployed on edge devices due to the fewer parameters requiring training.

Regarding the time cost of adaptation, although
back-propagation is required throughout the entire model in our method,
it is important to note that during the parameter update process, only 1\% of a ResNet-18’s parameters are updated.
This selective parameter update can save time during adaptation. Besides, the symmetry loss used in our work
can be fast calculated, unlike the one in RUDA that requires many times of image rotation. Here, we report the time cost
of adaptation (duration/iteration) including loss calculation or not. By comparing the time solely for model updates (without loss calculation), we are 3 times faster than RUDA. And on the total time cost for adaptation (with loss calculation), we achieve a speed 10 times faster than RUDA, let alone the slower PnP-GA.
The plausible results further demonstrate the importance of prompt tuning and our loss selection.

\begin{table}[!t]
\centering
\small
\scalebox{0.87}{
\begin{tabular}{|c|c|c|c|c|}
\toprule
\textbf{Method} &\textbf{$\mathcal{D}_E \rightarrow \mathcal{D}_M$} & \textbf{$\mathcal{D}_E \rightarrow \mathcal{D}_D$} & \textbf{$\mathcal{D}_G \rightarrow \mathcal{D}_M$} &\textbf{$\mathcal{D}_G \rightarrow \mathcal{D}_D$}\\ 
\midrule
\textbf{Baseline } & 8.02 & 7.30& 7.79&8.19\\
\midrule
\textbf{DAGEN }  & 7.53& 8.46&9.31 & 12.05\\
\textbf{GazeAdv }   & 8.48&7.70 &9.15 &11.15 \\
\textbf{Gaze360}  & 7.86& 9.64& 7.71& 9.54 \\
\midrule
\textbf{PnP-GA* } &6.91&7.18&7.36&8.17\\
\textbf{RUDA* }  & \underline{6.86}& \underline{6.84}&\underline{6.96} &\underline{5.32} \\
\textbf{TPGaze}   &\textbf{6.30} & \textbf{5.89} &\textbf{6.62} &\textbf{5.04} \\
\bottomrule
\end{tabular}}
\caption{Comparison with the state-of-the-art methods in terms of gaze error in degree, including source available UDA methods (middle rows) and our re-implementation (*) of source-free UDA methods under our personalization setting (bottom rows). The best results are in bold and the second best results are with underline. }
\label{tab:comp}
\end{table}

\begin{table}
\centering
\scalebox{0.8}{
\setlength{\tabcolsep}{3.2mm}{
\begin{tabular}{|c|c|c|c|c|}
\toprule
Method & Tunable Params &Time w/ Loss & Time w/o Loss   \\ \midrule
\textbf{PNP-GA} & 116.9M & 0.390s &0.359s \\ 
\textbf{RUDA}    & \underline{12.20M} & \underline{0.287s} & \underline{0.089s} \\ 
\textbf{TPGaze}    & \textbf{0.125M} & \textbf{0.029s} & \textbf{0.028s} \\ \bottomrule
\end{tabular}}}
\caption{Comparison in terms of adaptation efficiency (duration/iteration). Tunable parameters is the quantity of parameters that are updated during adaptation.``Time w/ Loss" and ``Time w/o Loss" respectively denote the time consumption in adaptation with or without loss calculation. }
\label{tab:efficiency}
\end{table} 

\subsection{Ablation Study}
We  here conduct
ablation studies to reveal the influence of the key components in our method. They are presented as follows:
(a) \textbf{Baseline}: using ResNet-18 without personalization. (b) \textbf{Update All}: updating all parameters of the network (meta-learned) for personalization instead of prompt tuning. (c) \textbf{No Meta}: using $\mathcal{L}_{per}$ for personalization with randomly initialized prompts instead of meta prompts. (d) \textbf{TPGaze}: our proposed method.

The quantitative results of the ablation studies are shown in Table  \ref{tab:ablation}, demonstrating that our method with all components works the best.
There are some other observations worth mentioning too.
First, by comparing \textbf{TPGaze} with \textbf{Update All}, we find that prompt tuning is surprisingly better than tuning all parameters. This substantiates the effectiveness of introducing prompt for personalization.  Second, by comparing \textbf{TPGaze} with \textbf{No Meta}, we validate the importance of using meta-learning for initializing prompts. Without meta-initialization, the performance of personalization drops. This highlights the significance of meta-learning in aligning the unsupervised loss with gaze error, leading to a more stable personalization process.
\begin{table}[!t]
\centering
\scalebox{0.8}{
\begin{tabular}{|c|c|c|c|c|}
\toprule
\textbf{Method} & $\mathcal{D}_E \rightarrow \mathcal{D}_M$ & $\mathcal{D}_E \rightarrow \mathcal{D}_D$ & $\mathcal{D}_G \rightarrow \mathcal{D}_M$ & $\mathcal{D}_G \rightarrow \mathcal{D}_D$ \\
\midrule
\textbf{Baseline} & 8.02 & 7.30& 7.79&8.19  \\ 
\textbf{Update All} & \underline{6.47} & \underline{6.10} & 6.71 & \underline{5.58}  \\
\textbf{No Meta} & 6.59 & 6.18 & \underline{6.68} & 5.76 \\ 
\textbf{TPGaze} & \textbf{6.30} & \textbf{5.89} & \textbf{6.62} & \textbf{5.04}\\
\bottomrule
\end{tabular}}
\caption{Ablation study of two components in our proposed method. Ours full solution (TPGaze) performs the best.}
\label{tab:ablation}
\end{table}


\subsection{Additional Analysis}
\subsubsection{Influence of Prompt Size}
To study the influence of altering the number of layers with tunable prompt,
we here conduct an additional analysis. The experimental results are shown in Table \ref{tab:prompt_size}. We gradually increase the number from 0 (no prompt) to 17 (all convolutional layers). 
As the prompt size increases,
the average gaze error first decreases until the 9-th layer.  Notably, the gaze errors when adding prompt to 9 layers and 13 layers are comparable, while the number of tunable parameters required for 13 layers is 48.9\% higher than that of 9 layers. To maximize the benefits of accurate and efficient inference, we suggest incorporating prompts into 9 convolutional layers.

It is worth noting that when we add prompt to all the layers (17 convolutional layers), the performance decreases significantly.  This could be attributed to the fact that the deepest prompt, located on the 17-th layer, can directly affect the output of the backbone.  Since we do not update the final linear layer during personalization, the layer maintains the original mapping from the original backbone's outputs to the gaze direction.  Thus, any major changes to the output features of the backbone may result in a defective mapping of the linear layer, leading to a catastrophic decrease in performance.  This highlights the need for careful consideration when tuning prompts with respect to deeper layers.

\begin{table}[!t]
\centering
\normalsize
\scalebox{0.73}{
\begin{tabular}{|c|c|c|c|c|c|}
\toprule
\textbf{\# Conv} & $\mathcal{D}_E \rightarrow \mathcal{D}_M$ & $\mathcal{D}_E \rightarrow \mathcal{D}_D$ & $\mathcal{D}_G \rightarrow \mathcal{D}_M$ & $\mathcal{D}_G \rightarrow \mathcal{D}_D$& \textbf{\# Param} \\
\midrule
0 & 8.02 & 7.30 & 7.79 & 8.19 & 0\\ 
1 & 6.48 & 6.48 & 6.91 & \underline{5.03} & 8.17K \\
5 & 6.38 & 6.29 & \underline{6.67} & \textbf{5.02} & 66.5K \\
 9 & \underline{6.30} & \underline{5.89} & \textbf{6.62} & 5.04 & 125.7K \\
 13 & \textbf{6.05} & \textbf{5.63} & 6.96 & 5.45 & 186.8K \\
 17 & 12.45 & 14.23 & 22.67 &  27.34& 251.1K\\
\bottomrule
\end{tabular}}
\caption{Influence of adding prompt to the different number of convolutional layers. ``\# Param" is the total number of parameters in the prompt. }
\label{tab:prompt_size}
\end{table}

\begin{table}[!t]
\centering
\scalebox{0.77}{
\begin{tabular}{|c|c|c|c|c|c|}
\toprule
\textbf{\# Samp} & \textbf{$\mathcal{D}_E \rightarrow \mathcal{D}_M$} & $\mathcal{D}_E \rightarrow \mathcal{D}_D$ & $\mathcal{D}_G \rightarrow \mathcal{D}_M$ & $\mathcal{D}_G \rightarrow \mathcal{D}_D$ & Avg. \\
\midrule
1 & 6.51 & 6.00 & \underline{6.54} &  5.21 &6.06 \\
5 & 6.30 &5.89 & 6.62 &  5.04 & 5.96 \\
10 & \underline{6.18} & \underline{5.51} & 6.67 &  \underline{5.03} & \underline{5.85}\\
15 & \textbf{6.14} & \textbf{5.26} & \textbf{6.54} &  \textbf{5.02} & \textbf{5.74}\\
\bottomrule
\end{tabular}}
\caption{The influence of using difference size of samples for personalization. We observe that in general more samples usually result in lower gaze error.}
\label{tab:data_size}
\end{table}

\subsubsection{Influence of Data Size}
As noted in Section \ref{sec:implementation}, we only adopt 5 samples for the personalization. Here, we reveal the influence of using a different number of unlabeled samples for personalization. The results are presented in Table \ref{tab:data_size}. Specifically, when we vary the number of samples from 1 to 15, we can observe that as the number
increases, the gaze errors tend to decrease,
fitting our intuition that more unlabeled samples can potentially help improve the accuracy of the personalized gaze estimation model.
\subsubsection{Visualization of Adaption Results}
\begin{figure}[!t]
\centering
\includegraphics[width=\linewidth]{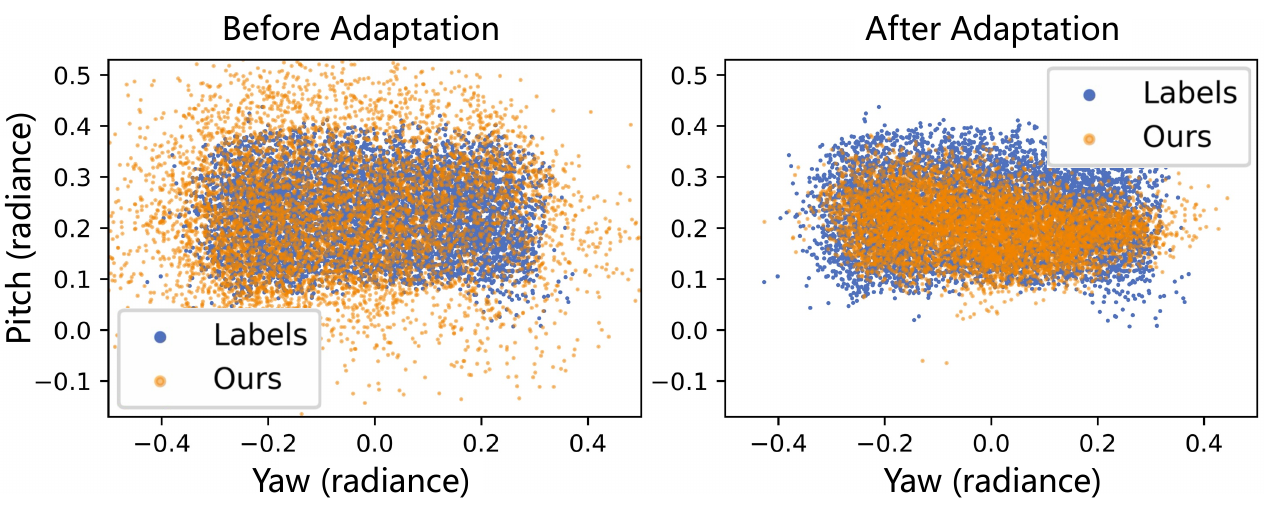}
\caption{Distribution of gaze estimation results and ground-truth labels before and after personalization. Results are the personalization from $\mathcal{D}_G $ to $ \mathcal{D}_D$.}
\label{fig:distribution}
\end{figure}

To visualize the personalization results, we show the distribution of gaze estimation results before and after adaptation in Figure \ref{fig:distribution}. It can be observed that the distribution of our approach (after adaptation) is considerably closer to the distribution of ground-truth labels, indicating a better adaptation performance. 

\subsubsection{Breakdown of the Personalization Results}
\begin{figure}[!t]
\centering
\includegraphics[width=0.8\linewidth]{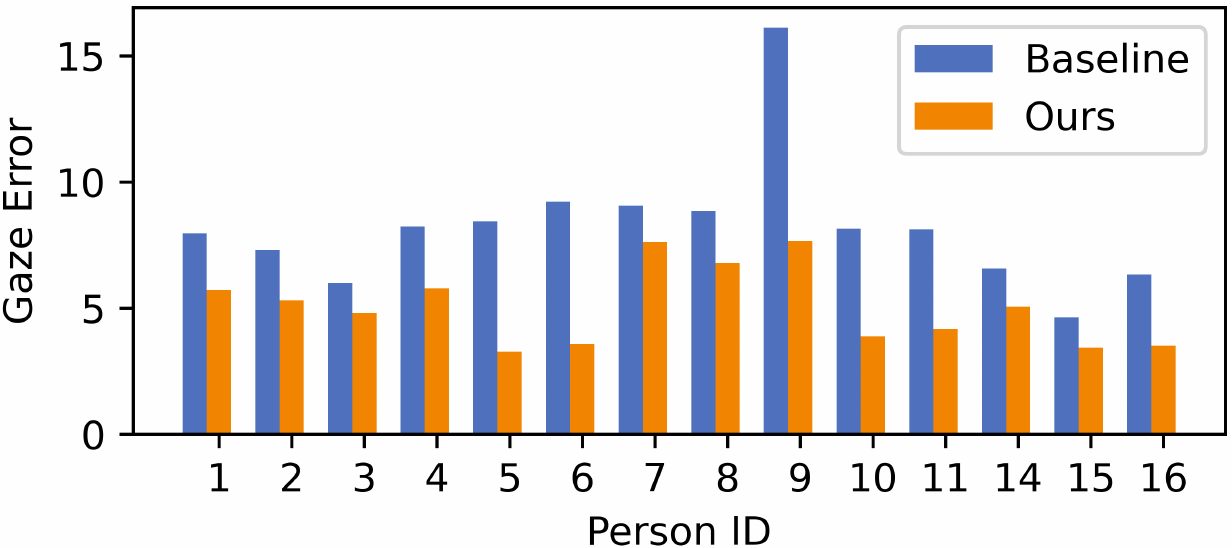}
\caption{Breakdown of personalization results from $\mathcal{D}_G $ to $ \mathcal{D}_D$.} 
\label{fig:breakdown}
\end{figure}

Due to the page limit, we primarily present the average gaze error across all individuals in a specific dataset. To provide further insight into our personalization results, we present a breakdown of the personalization results from $\mathcal{D}_G $ to $ \mathcal{D}_D$ in Figure \ref{fig:breakdown}.
It can be observed that our method can consistently outperform the baseline, as indicated by the lower gaze error. 

\section{Conclusion}
In this paper, we present an efficient and accurate method to personalize gaze estimation at test time, without relying on labeled data.  To achieve efficient personalization, we employ prompt tuning techniques. Moreover, we ensure that minimizing unsupervised loss aligns with minimizing gaze error through meta prompt. Our results show significant improvements against the state-of-the-art methods in terms of adaptation speed and accuracy.
Besides, extensive additional analysis further demonstrates the strong performance and desirable properties of our proposed approach.
\appendix
\section{Details of Loss Function}\label{sec:loss}
This section aims to offer additional details regarding the personalization loss (Equation 3 in our main body).

Personalization loss $\mathcal{L}_{per}$ is an unsupervised loss function used in the pre-training and personalization phases. Personalization loss function is indeed the \textbf{symmetry loss}, and is defined as follows:
\begin{equation}\label{eq:two_loss}
\begin{aligned}
    \mathcal{L}_{per}(f_\theta(\cdot), \cdot) &= \mathcal{L}_{sym}(f_\theta(\cdot), f_\theta(\mathcal{T}(\cdot))),
\end{aligned}
\end{equation}
\noindent
where $\mathcal{L}_{sym}$ is
the left-right symmetry loss, and $\mathcal{T}$ is the horizontal flip operation. $\cdot$ represents the potential image used to calculate the losses.

Symmetry loss, an unsupervised gaze loss, assists the model in making equivariant predictions on an image before and after a horizontal flip transformation. Given a normalized face image  $x_i^s$, we note its gaze label $G = (G_\Theta, G_\Phi)$, where $G_\Theta$ is the pitch angle and $G_\Phi$ is the yaw angle. When we apply horizontal flip on the face image $\mathcal{T}(x_i^s)$, the resulting gaze direction after the horizontal flip $G^\ast$ exhibits equivalent pitch $G_\Theta$ and negative yaw $-G_\Phi$ as compared to the original image. In favor of this left-right symmetry property, we adopt it as our second unsupervised loss. The left-right symmetry loss can be defined as:


\begin{equation}\label{eq:sym_loss}
\begin{aligned}
&\mathcal{L}_{sym}(f_\theta(x_i^s), f_\theta(\mathcal{T}(x_i^s))) = \\
&\frac{1}{2} \|f_\theta(x_i^s)- \begin{bmatrix}
         1 & 0\\ 
         0 & -1   
     \end{bmatrix} f_\theta(\mathcal{T}(x_i^s))\|_1,
\end{aligned}
\end{equation}
\noindent
where $ \begin{bmatrix}
         1 & 0\\ 
         0 & -1   
     \end{bmatrix}$ is the transformation matrix for gaze prediction, resulting in the same effect of adding a minus on the predicted yaw of the flipped image.

\section{Reproducing RUDA and PnP-GA}
\subsection{Reimplementation Details of RUDA}
We use the official code provided by the authors of RUDA \cite{RUDA} for our reimplementation, and only made slight modifications to the dataloaders to adapt to our dataset format. During the Rotation-Augmented Training (RAT) phase, we set the number of random rotations $K$ to 5 and the range of rotation to $[-90^{\circ}, 90^{\circ}]$, in alignment with RUDA \cite{RUDA}. Same as RUDA, the batch size is set to 80 and the model is trained for 10 epochs in the RAT phase. In the adaptation phase, we use the best epoch from the RAT phase for each cross-domain task. $K$ is set to 20 and the range of rotation remains at $[-90^{\circ}, 90^{\circ}]$. Under our personalization setting, since we only use the first 5 images of each person in the target domain for adaptation, we set the batch size as 5. We also train the model for 10 epochs in the adaptation phase. The momentum coefficient $\alpha$ in the EMA algorithm used by RUDA is set to 0.99. The Adam optimizer with a learning rate of
$10^{-4}$ and $\beta = (0.5, 0.95)$ is used. By using the same hyperparameters as RUDA \cite{RUDA} and only changing the adaptation setting to match our personalization setting, we ensure a faithful reproduction of RUDA.

\subsection{Reimplementation Details of PnP-GA}
Similarly, for the reimplementation of PnP-GA, we use the official open-source code released by the authors of PnP-GA \cite{pnp-gaze}. As described by \cite{pnp-gaze}, we employ a group of 10 pre-trained models to use for adaptation. Since the pre-trained models used by the original PnP-GA paper are not available to us, we use 10 epochs from our pre-training stage as 10 models. In domain adaptation, we also use the first 5 images of each person in the target domain, with a batch size of 5. Since the original PnP-GA can be converted to a source-free domain adaptation method by removing the source domain's $\mathcal{L}_1$ loss (sg), we do so to align with our source-free personalization setting. We keep the rest of the settings identical to PnP-GA \cite{pnp-gaze}. The significance level $\epsilon$ and weight parameter $\gamma$ associated with the outlier-guided loss are set to 0.05 and 0.01, respectively. Hyperparameters $\lambda_1$ and $\lambda_2$ are associated with the Jensen-Shannon divergence and outlier-guided loss components in the total loss equation. $\lambda_1 = 0.01$ and $\lambda_2 = 0.1$, same as PnP-GA \cite{pnp-gaze}. The momentum coefficient $\alpha$ is 0.99. The Adam optimizer with a learning rate of $10^{-4}$ is used.

\bibliography{aaai24}

\end{document}